\documentclass[conference,10pt]{IEEEtran}
\IEEEoverridecommandlockouts

\usepackage{cite}
\usepackage{amsmath,amssymb,amsfonts}
\usepackage{algorithmic}
\usepackage{subfigure}
\usepackage{mathtools}
\usepackage{graphicx}
\usepackage{textcomp}
\usepackage{xcolor}
\def\BibTeX{{\rm B\kern-.05em{\sc i\kern-.025em b}\kern-.08em
    T\kern-.1667em\lower.7ex\hbox{E}\kern-.125emX}}

\begin{document}

\title{Adversary Helps: Gradient-based Device-Free Domain-Independent Gesture Recognition\\}

\author{\IEEEauthorblockN{1\textsuperscript{st} Jianwei Liu}
\IEEEauthorblockA{\textit{} 
\textit{Zhejiang University}\\
Hangzhou, China \\
liujianwei@stu.xjtu.edu.cn}
\and
\IEEEauthorblockN{2\textsuperscript{nd} Jinsong Han}
\IEEEauthorblockA{\textit{} 
\textit{Zhejiang University}\\
Hangzhou, China \\
hanjinsong@zju.edu.cn}
\and
\IEEEauthorblockN{3\textsuperscript{rd} Feng Lin}
\IEEEauthorblockA{\textit{} 
\textit{Zhejiang University}\\
Hangzhou, China \\
flin@zju.edu.cn}
\and
\IEEEauthorblockN{4\textsuperscript{th} Kui Ren}
\IEEEauthorblockA{\textit{} 
\textit{Zhejiang University}\\
Hangzhou, China \\
kuiren@zju.edu.cn}
}

\maketitle

\begin{abstract}
Wireless signal-based gesture recognition has promoted the
developments of VR game, smart home, etc. However, traditional approaches suffer from the influence of the domain gap. Low recognition accuracy occurs when the recognition model is trained in one domain but is used in another domain. Though some solutions, such as adversarial learning, transfer learning and body-coordinate velocity profile, have been proposed to achieve cross-domain recognition, these solutions more or less have flaws. 
\par In this paper, we define the concept of domain gap and then propose a more promising solution, namely DI, to eliminate domain gap and further achieve domain-independent gesture recognition. DI leverages the sign map of the gradient map as the domain gap eliminator to improve the recognition accuracy. We conduct experiments with ten domains and ten gestures. The experiment results show that DI can achieve the recognition accuracies of 87.13\%, 90.12\% and 94.45\% on KNN, SVM and CNN, which outperforms existing solutions.
\end{abstract}


\section{introduction}
Human gesture recognition plays a core role in human computer interaction, meanwhile, it enables many human-centered applications such as patient surveillance, smart home and virtual reality gam. Though traditional approaches have outstanding gesture recognition effectiveness, they are defective more or less. For example, camera-based approaches \cite{DBLP:conf/cvpr/GkioxariGDH18,DBLP:journals/sigmobile/LiLZ16} have to obtain the visual body information of the user, which may causes the invasion of visual privacy. Sonar-based approaches \cite{DBLP:conf/icassp/KalgaonkarR09,DBLP:journals/imwut/NandakumarTKG17} would not make users under visual surveillance, however, they require high-cost and cumbersome hardware. Wearable sensor-based approaches \cite{DBLP:journals/imwut/GuanP17,DBLP:journals/csur/BullingBS14}, though are light and visual information-needless, they face the flaws of limit of sensing range and relatively bad user-friendness.

\par Later on, with the rapid development of the wireless sensing technique, a more promising and attractive solution becomes WiFi signal-based non-intrusive gesture recognition. For instance, E-eyes \cite{DBLP:conf/mobicom/WangLCG0L14} is the pioneer to utilize WiFi signal as gesture sensing media, which is followed by massive similar attempts such as WiGest \cite{DBLP:conf/infocom/Abdel-NasserYH15}, WIMU \cite{DBLP:conf/mobisys/VenkatnarayanPS18} and CSI-Net \cite{wang2018csi}. However, traditional solutions only focus on the high recognition accuracy under single domain but neglect the impact of domain gap. If there are multiple domains involved in the recognition requirement, the recognition effectiveness would drop continuously with the increase of domains. The reason behind is that signals are not only distorted by gestures but also changed by domains. Thus, there are two gaps in the multi-domain data distribution caused by gesture difference and domain difference: gesture gap and domain gap. Traditional solutions work by utilizing the gesture gap as the boundary between gestures, yet the domain gap would make the boundary ambiguous and further reduce the recognition accuracy.

\par Recently, several works are dedicated to cross-domain gesture recognition. EI \cite{DBLP:conf/mobicom/JiangMMYWYXSMKX18} extracts environment-independent features by using an adversarial network. CrossSense
\cite{DBLP:journals/cea/RighiGKDC20} utilizes transfer learning and ensemble learning to alleviate the impact of domains. WiAG \cite{DBLP:conf/mobisys/VirmaniS17} learns a translation function to transform data between domains. Widar3.0 \cite{DBLP:conf/mobisys/ZhengZ0ZLW019} extracts the body-coordinate velocity profile to achieve domain-independent gesture recognition. Nevertheless, these works have obvious flaws. EI adds the constraint loss of activity percentage on the classifier, but the activity percentages may be different in different scenarios. The learning models used in CrossSense are ponderous and lack of computing-friendness. Widar3.0 is incapable to recognize static gestures because they have no velocity component. WiAG needs to generate virtual samples for all possible domains and gestures, which is not computing-friendly when domain number is large. Meanwhile, the recognition accuracies of EI and CrossSense are relatively low.

\par In this paper, we propose DI, which does not have above flaws but can achieve comparable recognition accuracy with Widar3.0 and WiAG. Specifically, DI utilizes a novel deep model, namely AH-Net, to convert domain-specific samples into domain-independent sample. AH-Net achieve conversion goal by leveraging the technique of fast gradient sign method-based adversarial sample to eliminate the domain gap. After conversion, k-nearest neighbours (KNN), support vector machine (SVM) and convolutional neural network (CNN) are all options which can be used as gesture recognizer.

\par We evaluate DI with ten gestures and ten domains. The experiment results show that DI can reach the accuracy of 87.13\%, 90.12\% and 94.45\% on KNN, SVM and CNN, which outperforms previous works. In summary, our contributions are listed as followings:
\begin{itemize}
    \item We proposes a novel framework named DI to achieve domain-independent gesture recognition.
    \item We propose a novel deep model, namely AH-Net, to convert domain specific samples into domain-independent samples.
    \item We conducted experiments with ten gestures and ten domains, the results show that DI can achieve the accuracy of 94\%+, which outperforms existing works.
\end{itemize}

\section{Related work}
\textbf{Learning-based RF technique:} With the development of deep learning, massive works have utilized deep model to solve tough problems in RF sensing. CSI-Net \cite{wang2018csi} is a novel convolutional neural network (CNN)-based deep model specifically designed for WiFi-based gesture recognition. Person-in-wifi \cite{DBLP:conf/iccv/0037ZPHH19} devises a U-net based deep model to estimate fine-grained human skeletons and gestures. Wang \textit{et al.} \cite{DBLP:conf/infocom/WangLCLXWHL18} leverage CNN to localize the in-air input written in front of a tag array. E-eyes \cite{DBLP:conf/mobicom/WangLCG0L14} employs commercial WiFi system and K-nearest neighbours algorithm to conduct activity recognition. wang \cite{DBLP:journals/jsac/WangLSLL17} use HMM as learning model to identify activities, the input feature is the power distribution extracted from Doppler frequency shift components. Wisee \cite{DBLP:conf/mobicom/PuGGP13} also uses Doppler frequency shift to achieve whole-home and device-free activity recognition. 

\par Different from above works, DI is the pioneer that proves the avalibility of adversarial sample technique in RF field. Meanwhile, DI not only focus on recognition accuracy but also the cross-domain ability.

\par \textbf{Domain-independent RF techniques:} Recently, researchers have noticed the drawback of the diversity caused by multi-domain settings and proposed several solutions. Butterfly \cite{DBLP:journals/imwut/HanQYWDLR18} overcomes the diversity of environments in RFID identifications by using a pair of tags which nearly experience the same environment. Virmani \cite{DBLP:conf/mobisys/VirmaniS17} shows that one can leverage a translation function to map signals to target domain features. In order to obtain the ability of activity recognition in multiple environments, EI \cite{DBLP:conf/mobicom/JiangMMYWYXSMKX18} utilizes an adversarial network to extract environment-independent features from original signals. CrossSense \cite{DBLP:journals/cea/RighiGKDC20} simultaneously utilize transfer-learning and ensemble learning to achieve cross-site activity recognition. Zheng \textit{et al.} \cite{DBLP:conf/mobisys/ZhengZ0ZLW019} first propose the concept of body-coordinate velocity profile (BVP) and extract BVP from Doppler frequency shift to cross-domain recognize gestures. 

\par Different from EI. DI does not add any loss constraints on activity number rate. Moreover, the multiple algoritms in CrossSense makes it less explainable and cost much time during training, yet DI only employs one model. BVP is an effective way for dynamic gesture recognition. However, it can not work on static gestures like DI can do.

\section{Motivation and Preliminary}
In this section, we first introduce the impact of domain in RF-based activity/gesture recognition. Then we briefly introduce the principle and the application of FGSM.

\subsection{The impact of domain}
\begin{figure}[t]
    \centering
    \includegraphics[scale=0.5,trim=30 30 30 30,clip]{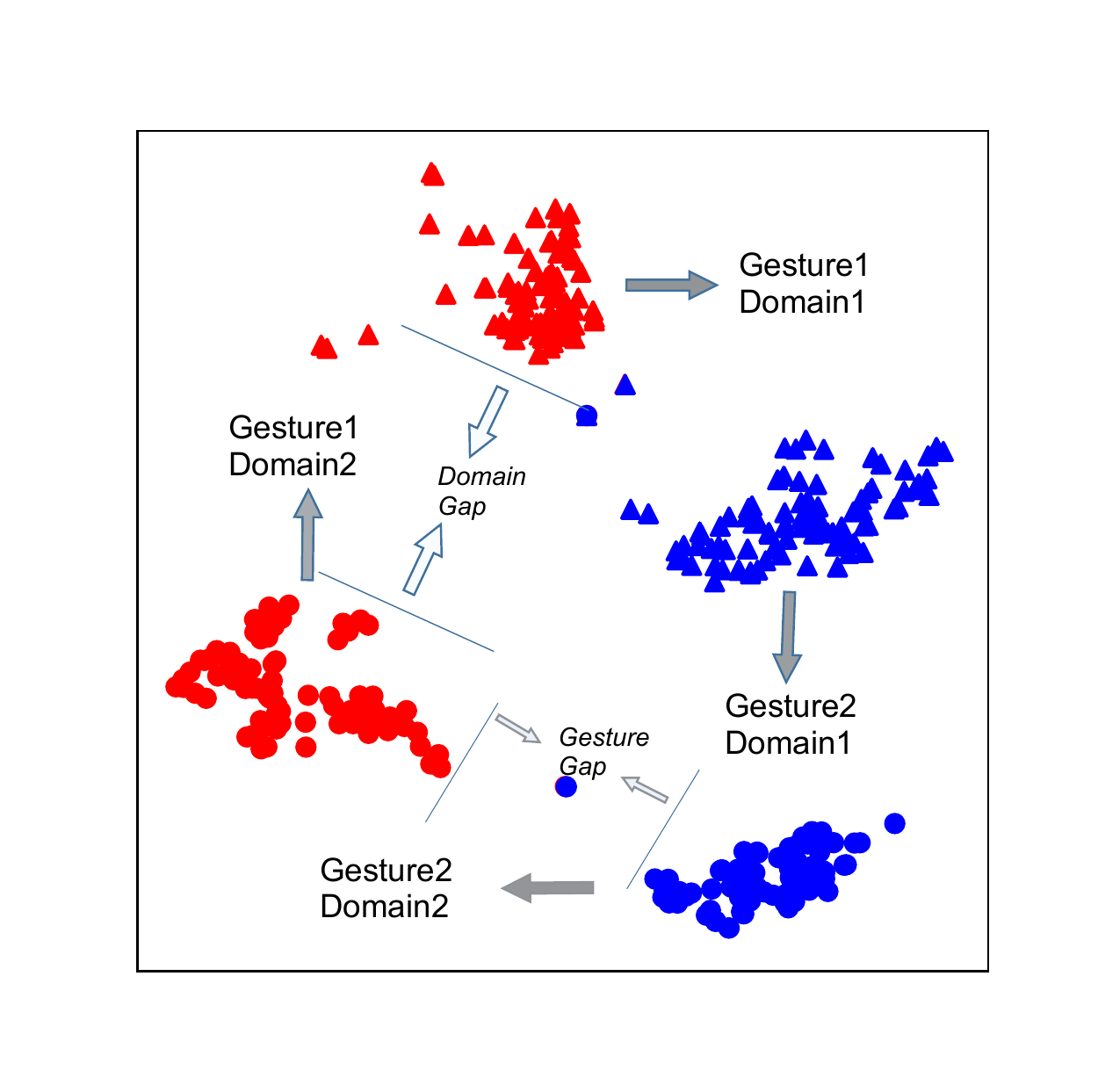}
    \caption{The signal sample is separated by both gesture gap and domain gap..}
    \label{fig:domain_gap}
\end{figure}

Though RF-based activity/gesture recognition has been pervasively studied in recent years, the practical use and landing are not well promoted. The reason behind is the diversity of the implementation domain, including person domain and environment. Since each specific domain has its own unique signal embedding scheme, in different domains, even the same gesture would yield different signal distortions. In this case, two signal measures of the same gesture, which are collected in two different domains, have a domain gap between each other. If those two signals are collected under not only different gesture settings but also different domian settings, there will be two categories of gaps between them: domain gap and gesture gap. As illustrated in Fig. \ref{fig:domain_gap}, we plot some signal samples after dimension reduction by t-SNE \cite{DBLP:journals/gandc/HorrocksHWWF19}. Different colors represent different gestures and different patterns represent different domains. Apparently, signal samples which belong to two gestures are divided into four clusters according to both the diversity of gestures and the diversity of domains. However, normal learning algorithm can only learn the gap between gestures yet neglect the gap between. Thus, normal learning algorithm can not deal with multi-domain signals. Motivated by this issue, we propose DI to eliminate the gap between domain by adding domain-eliminating factors. In this way, even traditional machine learning algorithms are qualified to recognize activities/gestures domain-independently.

\subsection{Fast Gradient Sign Method}
Adversarial sample attack falls into two categories according to whether the attacker has knowledge about the target model architecture or not: whitebox attack and blackbox attack. Fast gradient sign method (FGSM) \cite{DBLP:journals/corr/GoodfellowSS14} is one of the classic whitebox adversarial sample generation methods. Basically, adding perturbation on normal sample is the core of FGSM. This perturbation is calculated based on the signs of gradients. Specifically, the procedure to generate a adversarial sample by using FGSM can be segmented into three steps: 1) Training a normal deep model $M$ which can correctly classify sample $x$. 2) Calculating the gradient map of $x$ by back propagating on $M$ and obtaining the sign map $s$ of this gradient map. 3) Determining a factor $\epsilon$ and calculating $x^a$ by:
\begin{equation}
    x^a = x + \epsilon \times s.
\end{equation}
If $x^a$ is incorrectly classified by $M$, $X^a$ is qualified as an adversarial sample towards model $M$. 

\par Interestingly, in DI, the sign map of the gradient map is leveraged as a domain gap eliminator (DGE) rather than a deep model adversarial sample generator. 

\section{DI overview}
In this section we first briefly introduce the workflow of DI and then each module of DI is elaborated.

\par
\begin{figure}
    \centering
    \includegraphics[scale=0.6,trim=0 200 400 0,clip]{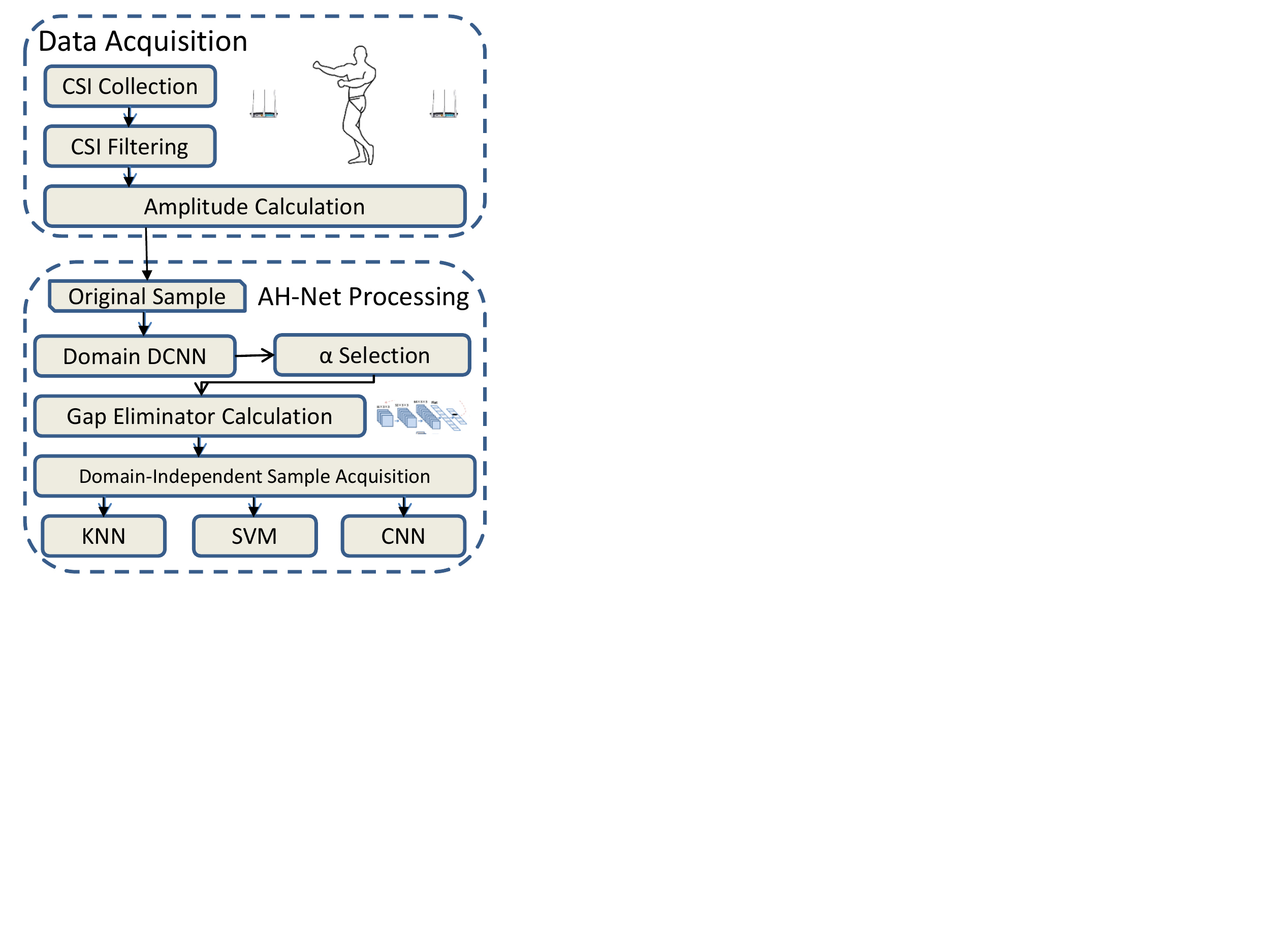}
    \caption{The workflow of DI. It consists of two modules: \textit{data acquisition} and \textit{AH-Net processing}.}
    \label{fig:workflow}
\end{figure}

 As shown in Fig. \ref{fig:workflow}, DI contains two modules: \textit{data acquisition} and \textit{AH-Net processing}. \textit{Data acquisition} is utilized to collect raw CSI data and conduct preprocessing of donoising. The output of \textit{data acquisition}, denoised amplitude, is the input of \textit{AH-Net processing}. \textit{AH-Net processing} aims to generate DGE for each original domain-specific sample. Afterwards, traditional machine learning algorithm and deep learning model are all optional gesture recognizer. The details are elaborated as followings:
\begin{itemize}
    \item \textbf{Data acquisition}: This module is the first module of DI. Due to that DI is a learning-based system, training set collection is the first step of \textit{data acquisition}. In this step, signals are collected when the user is acting a gesture between the transmitter and the receiver. Then signals are transformed as CSI data by using CSI tool \cite{DBLP:journals/ccr/HalperinHSW11}. Afterwards, CSI data is denoised through filtering. Kalman filter \cite{DBLP:journals/sensors/KangSZZG20} is an outstanding candidate in our empirical study. Further, DI calculates amplitude for each CSI sample and inputs amplitude samples into next module.
    \item \textbf{AH-Net processing:} In this module DI mainly trains the core of DI, namely AH-Net, and generates DGE for each sample. Specifically, DI first accepts the amplitude samples which are outputted from \textit{data acquisition} module as training set. Since AH-Net is mainly composed of a DCNN, i.e., domain DCNN, DI sufficiently trains the DCNN by using the training set. Thereafter, DI achieves the ability of DGE generation for any sample. Actually, AH-Net is the tool of DI to calculate DGE. In case each sample is added by its corresponding DGE, each sample becomes domain-independent. Afterwards, DI can select one of the learning algorithms from KNN, SVM and CNN as final recognizer. Comparing with KNN and SVM, CNN can reach higher recognition accuracy. However, KNN and SVM  are more lightweight. The final recognizer can be determined according to the configuration of the application scenario.   
\end{itemize}

\begin{figure*}
    \centering
    \includegraphics[scale=0.64,trim=0 380 0 20,clip]{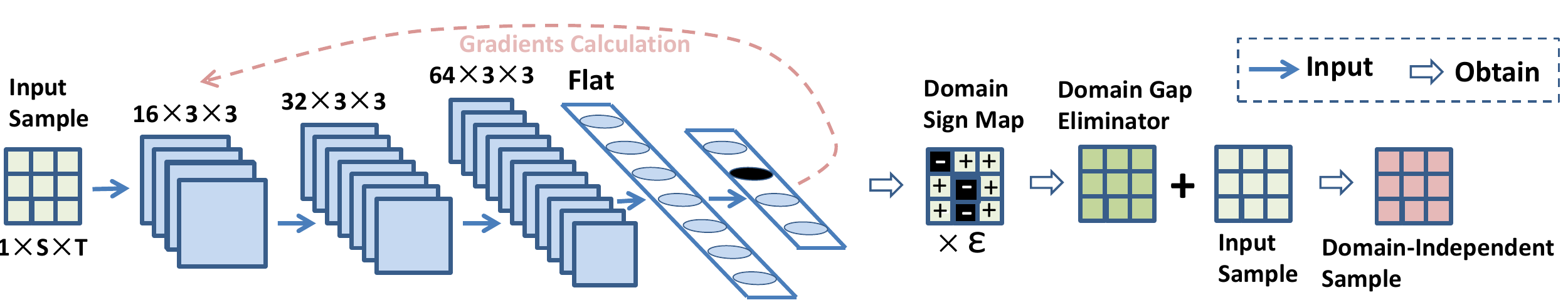}
    \caption{The architecture of AH-Net. Two DCNN are trained to generate domain gap eliminator.}
    \label{fig:AH_Net}
\end{figure*}

\section{Methodology}
This section elaborates the training set construction method in the first part. Then the architecture of AH-Net is introduced in detail. At last, we introduce the optimization objective function and training method. We arrange the details of DI usage in the last part.

\subsection{Training set construction}
Training set is of importance for a learning-based algorithm. The training sample of DI is special because domain is also a recognition target except for gesture. Specifically, each training sample has two labels: domain label and gesture label. When collecting training samples, each gesture is posing in every domain for several times. If there are 10 gestures and 10 domains, training set should contains $10 \times 10$ categories of training samples. After marking each sample with two labels, training set is constructed and utilized for domain DCNN training and gesture recognizer learning. 

\subsection{AH-Net architecture}
As shown in Fig. \ref{fig:AH_Net}, AH-Net mainly has a domain deep convolutional neural network and a procedure of domain-independent sample generation. Domain DCNN has three convolutional layers and two fully connected layers. Each convolutional layer is followed by a batch normalization function and a ReLU activation function. The kernel sizes of the first, the second and the third convolutional layer are $16 \times 3 \times 3$, $32 \times 3 \times 3$ and $64 \times 3 \times 3$
respectively. Each fully connected layer is followed by a Sigmoid activation function to increase nonlinearity.

\par The procedure of domain-independent sample generation is composed of three steps. First, for a sample $x$, DI obtains sign map for input $x$ through backpropagation in well-trained domain DCNN. Then the sign map is multiplied by a hyper-parameter $\alpha$, which generates a DGE for $x$. At last, $x$ is added by the DGE and becomes a domain-independent sample.

\subsection{Objective function and training method}
\textbf{Objective function:} For domain DCNN, DI utilizes cross entropy loss to optimize it, which can be formulated by:
\begin{equation}
  {LOSS}_d = -\sum_\text{c=1}^N y_c \log(P_c). 
  \label{eq:loss_d}
\end{equation}
In this formula, $N$ is the number of domains. $y_c$ is the indication variable and $P_c$ denotes the probability that the input is classified as domain $c$. If the the user chose CNN rather than KNN and SVM as the final gesture recognizer, the optimization objective function of gesture CNN is cross entropy loss as well, which is represented by:
\begin{equation}
  {LOSS}_g = -\sum_\text{c=1}^M y_c \log(P_c),  
\end{equation}
where $M$ is the number of gestures. $y_c$ and $P_c$ have similar representations with them in Eq. \ref{eq:loss_d}.

\par \textbf{Training method:} 
There are two parts of learning -based model need to be trainined in DI. DI first trains the domain DCNN with training samples and their domain labels by optimizing according to above mentioned loss function. Afterwards, DI needs to select an algorithm from KNN, SVM and CNN to recognize gestures. If CNN is the final chose, it would be trained via above mentioned loss function as well.

\subsection{DI usage}
For training samples, they have two labels: domain labels and gesture labels. DI uses training samples and domain labels to train AH-Net. Then gradient map for each training sample is calculated through backpropagation. Afterwards, DI obtains sign map for each gradient map. Further, each sign map is multiplied by well-selected $\alpha$ and becomes a DGE. At last, each training sample is added by its own DGE and becomes a domain-independent sample. DI converts all training samples into domain-independent training samples (DITS) and trains the gesture recognizer with DITS.

\begin{figure}
    \centering
    \includegraphics[scale=0.4,trim=0 10 20 20,clip]{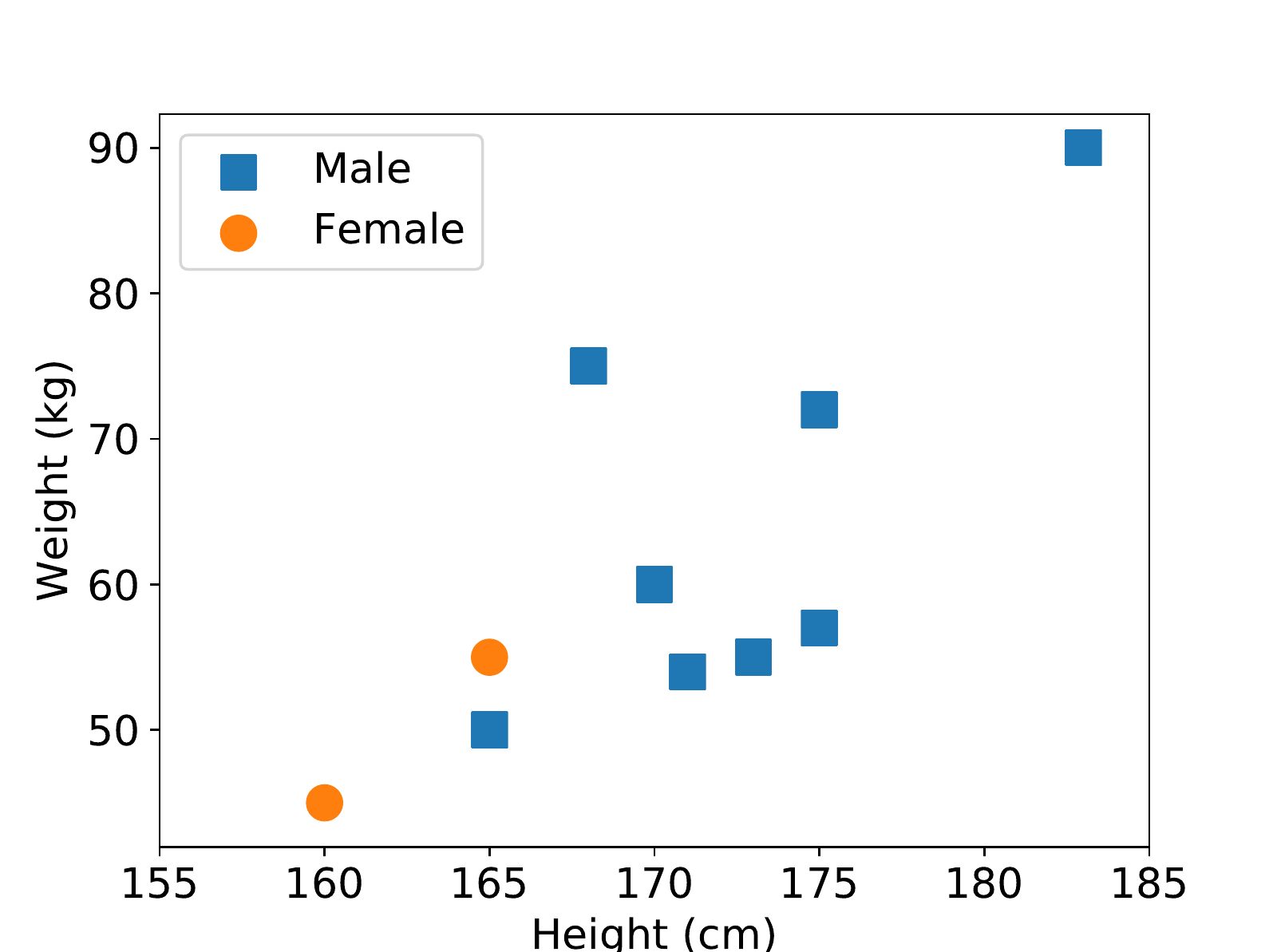}
    \caption{The gender, hight and weight of volunteers.}
    \label{fig:information}
\end{figure}

\begin{figure}
    \centering
    \includegraphics[scale= 0.45, trim=20 220 280 50,clip]{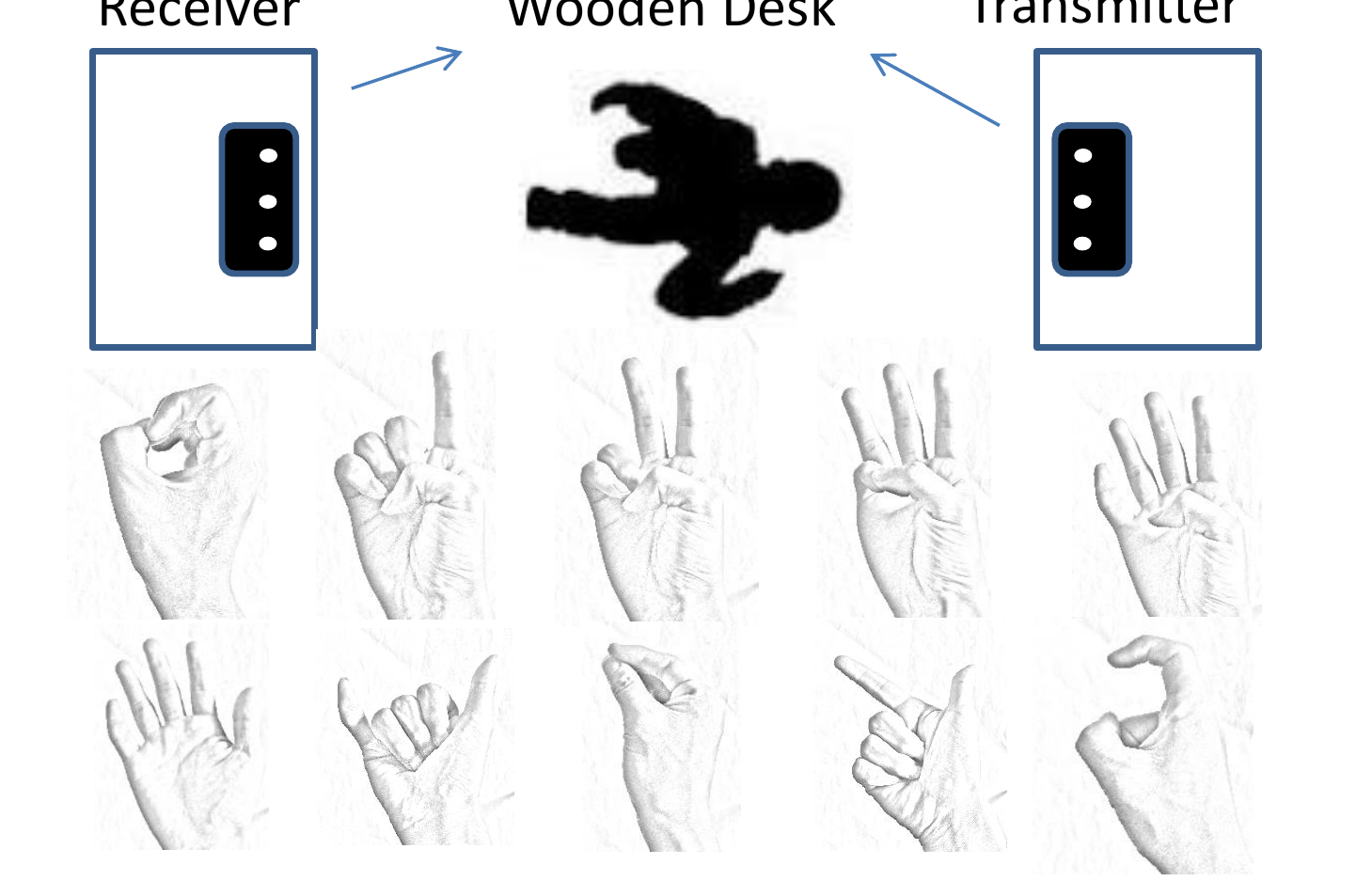}
    \caption{Experiment setup.}
    \label{fig:experiment_setupl}
\end{figure}

\begin{figure}[b]
    \centering
    \includegraphics[scale= 0.5, trim=20 20 20 40,clip]{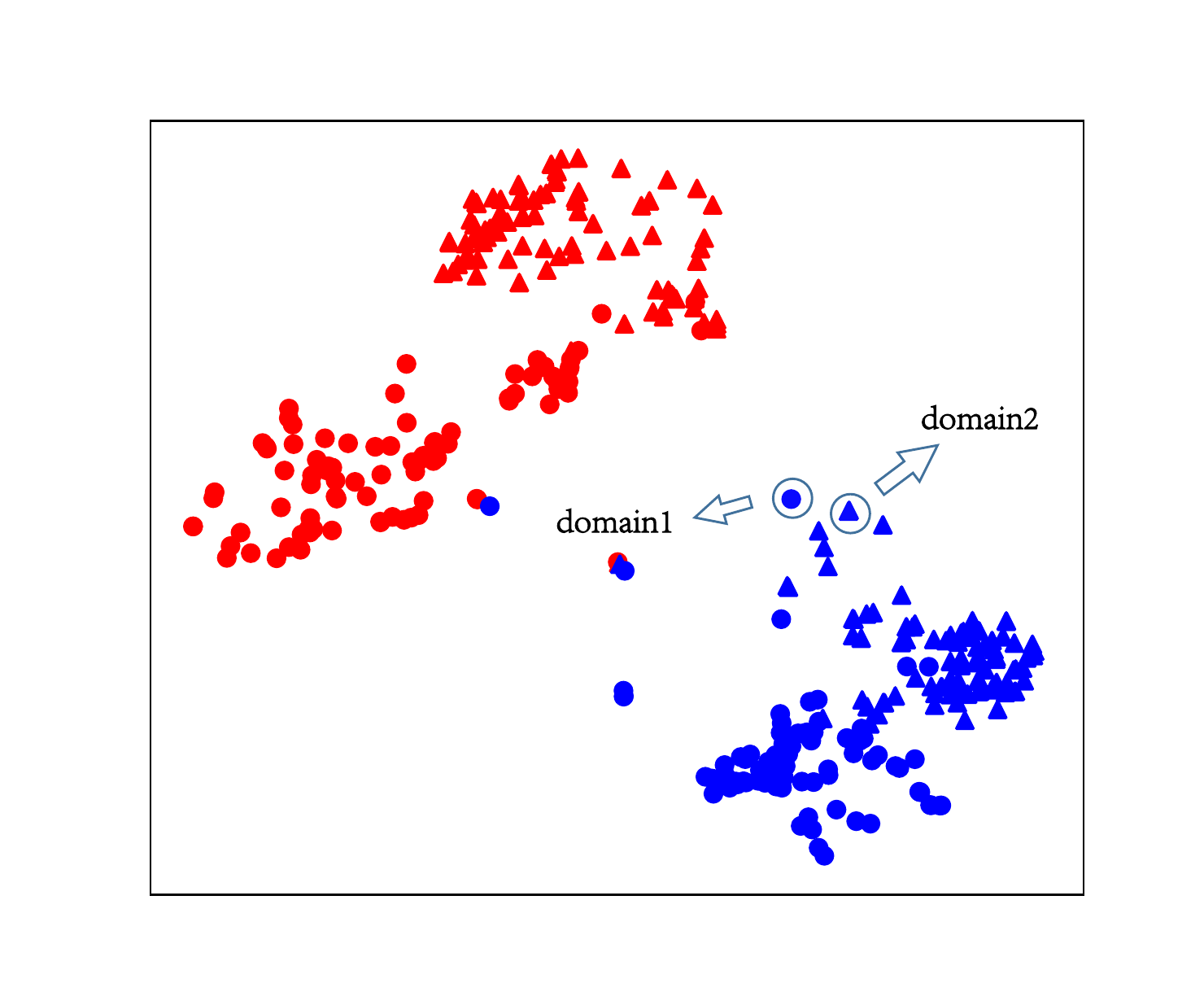}
    \caption{The domain gap is effectively eliminated.}
    \label{fig:domain_scatter}
\end{figure}

\par When DI receives a recognition request with a CSI sample, it first processes the CSI sample and obtains amplitude sample $x$ using \textit{data acquisition} module. Afterwards, $x$ is fed into domain DCNN and its domain $y$ is obtained. Thereafter, DI calculates $x$'s gradient map $g$ and $g$'s sign map $s$ with the formula:
\begin{equation}
    s = sign(\bigtriangledown_xf(\Theta, x, y)),
\end{equation}
in which $sign(\cdot)$ means obtaining the sign for each element. $\bigtriangledown_x$ means calculating gradient map for $x$. $f(\Theta,\cdot,\cdot)$ represents domain DCNN. Further, a domain-independent $x_{DI}$ is calculated by:
\begin{equation}
    x_{DI} = x + \alpha \times sign(\bigtriangledown_xf(\Theta, x, y)).
\end{equation}
At last, DI inputs $x_{DI}$ into gesture recognizer and obtain corresponding gesture index.

\begin{figure*}[t] 
\setlength{\belowcaptionskip}{-0.5cm}
\centering    
\subfigure[DI improves the recognition accuracy.] {
%  \label{a}     
\includegraphics[scale=0.365, trim =18 0 55 0 ,clip]{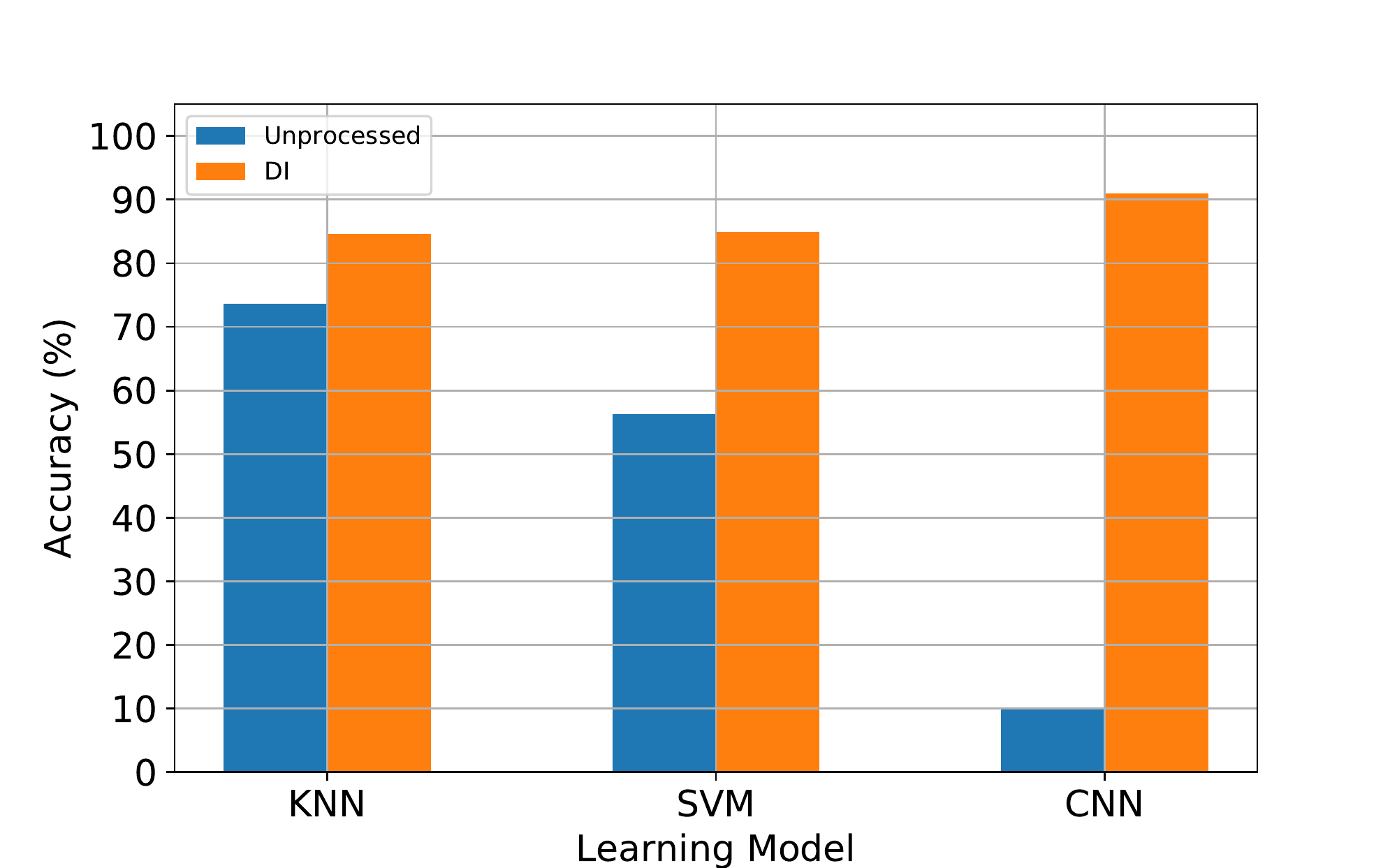}  
}
\hspace{10pt}
\centering
\subfigure[The effect of $\alpha$ on CNN.] { 
% \label{fig:b}     
\includegraphics[scale=0.365,trim=70 0 95 0,clip]{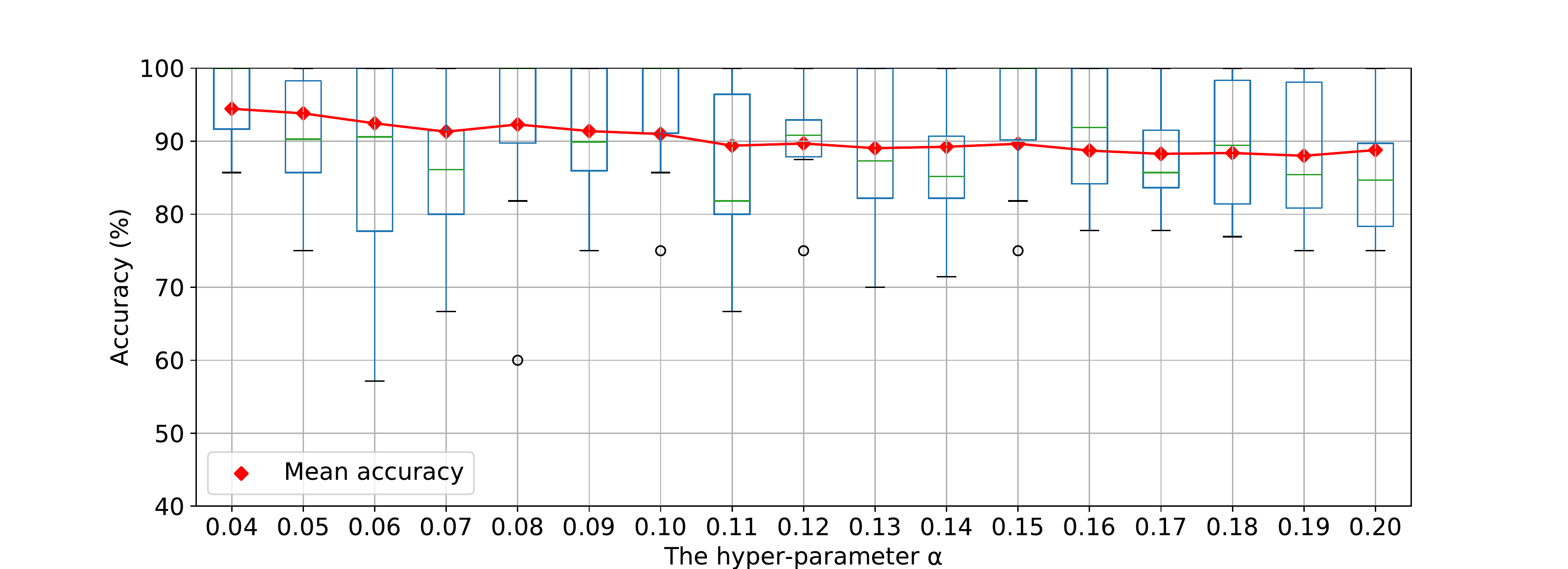}
}
\caption{(a) The validity of DI and (b) the effect of $\alpha$.}     
\label{fig:performance1}     
\end{figure*}

\begin{figure*}[t] 
\setlength{\belowcaptionskip}{-0.5cm}
\centering    
\subfigure[The effect of $\alpha$ on KNN.] {
%  \label{fig:a}     
\includegraphics[scale=0.30, trim =70 0 95 35,clip]{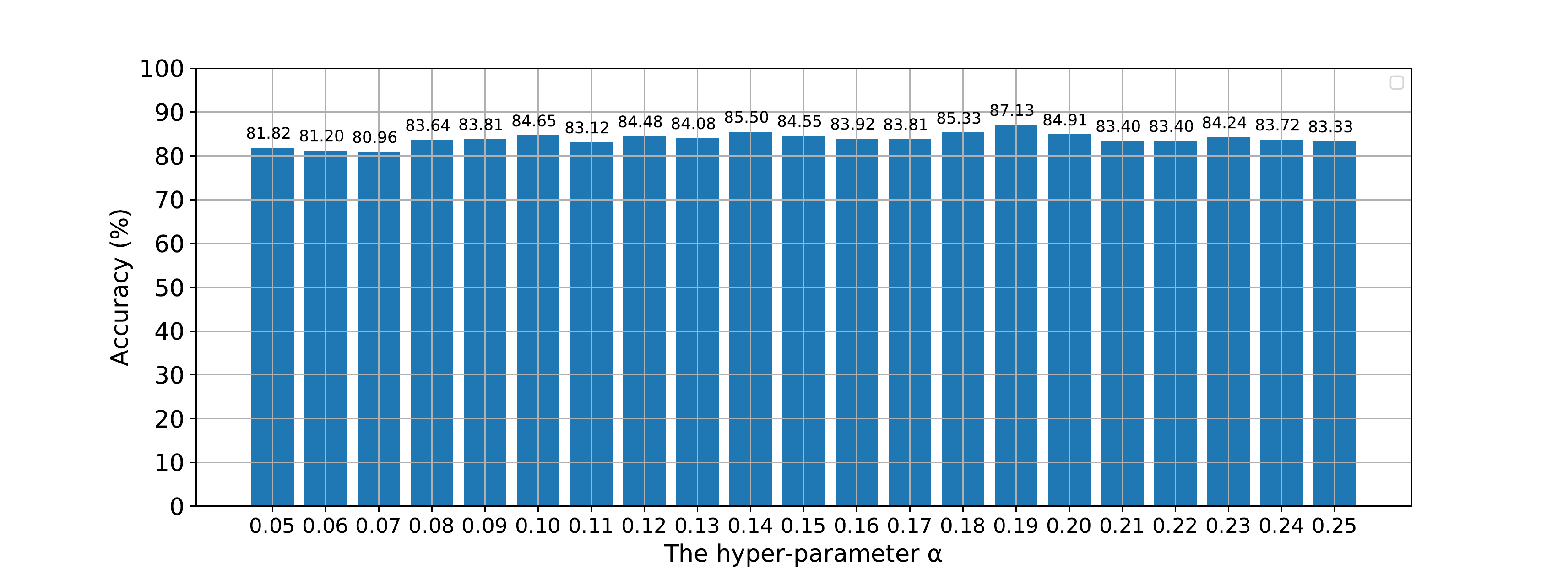}  
}
% \hspace{10pt}
\centering
\subfigure[The effect of $\alpha$ on SVM.] { 
% \label{fig:b}     
\includegraphics[scale=0.30,trim=70 0 95 35,clip]{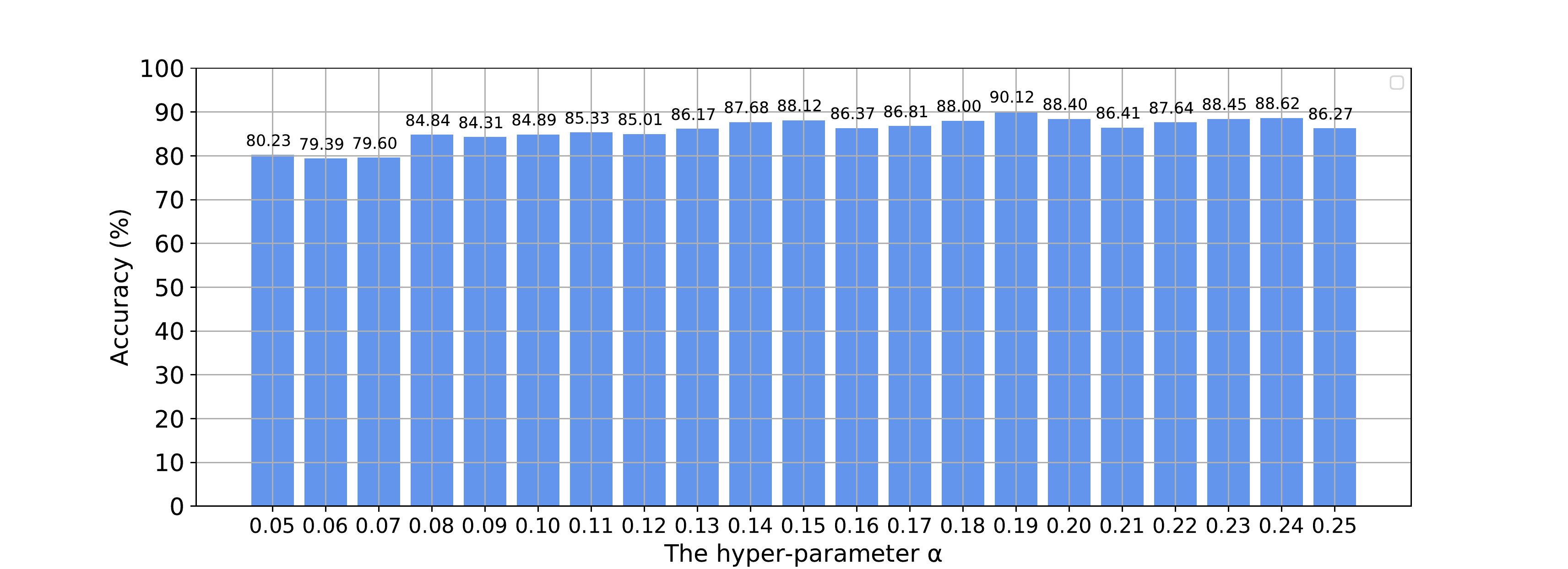}
}
\caption{The effects of $\alpha$ on (a) KNN and (b) SVM.}     
\label{fig:performance2}     
\end{figure*}

\section{Implementation and evaluation}
We conducted experiments with a commodity WiFi system with 10 volunteers, i.e., 10 domains, and 10 gestures. The information of 10 volunteers are displayed in Fig. \ref{fig:information}. The scatter plot of domains are dicentralized, which proves the validity of the domain. The hyper-parameter $\alpha$ is set as 0.1 by default.

\par \textbf{Hardware:} Both the transmitter and the receiver we used have three antennas. The receiver is a commercial WiFi router, the type of which is TPLink WDR7500-V3. Both the transmitter and the receiver are equipped with Atheros AR9500 Chips. We employ a computer with a i7-8700k CPU to collect signals.

\par \textbf{Software: }The operation system for signal collection is \textit{Ubuntu 14.04}. We use the Linux CSI tool provided by \cite{DBLP:journals/ccr/HalperinHSW11} to obtain CSI measurement. The collected CSI is then processed by \textit{Matlab} and \textit{Eclipse}. Our deep learning model is programmed by \textit{Python} language and \textit{Pytorch} framework. We run the model process on \textit{CUDA8.0}. 

\par \textbf{Experiment setup:} As shown in the top part of Fig. \ref{fig:experiment_setupl}, the volunteer is posing a gesture between the transmitter and the receiver. Both the transmitter and the receiver are 60cm away from the volunteer and 80cm off the ground. Ten gestures from zero to nine are displayed in the bottom part of Fig. \ref{fig:experiment_setupl}. 
 
\par \textbf{Evaluation metric:} The metric that commonly used for gesture recognition system evaluation is the accuracy, which can be formulated as:
\begin{equation}
    accuracy = \dfrac{N_{cor}}{N_{all}}.
\end{equation}
$N_{cor}$ is the number of correctly recognized test samples and $N_{all}$ is the number of all test samples.

\subsection{Evaluation with default $\alpha$}
Recalling that DI can convert each domain-specific sample into domain-independent sample, we use t-SNE \cite{DBLP:journals/gandc/HorrocksHWWF19} to reduce the dimension of four categories of domain-independent samples and plot them in Fig. \ref{fig:domain_scatter}. Different colors represent different gestures and different shapes represent different domains. Apparently, the gap of domain is effectively eliminated.

\par In order to show the superiority of domain-independent samples, we separately train KNN, SVM and CNN and then display the recognition accuracy in Fig. \ref{fig:performance1}(a). Before DI processing, the accuracy of KNN, SVM and CNN are 73.63\%, 56.32\% and 9.89\%, which are significantly low for a gesture recognition system. After DI processing, the accuracy of KNN, SVM and CNN increase to 84.65\%, 84.89\% and 90.92\%, respectively. The noticeable increase of the accuracy indicates that DI is efficient at domain gap eliminating. 

\subsection{Effect of $\alpha$ on CNN}
In order to explore the effect of the hyper-parameter $\alpha$ on CNN, we varied $\alpha$ from 0.04 to 0.20 with a stride of 0.01. The experiment results are shown i Fig. \ref{fig:performance1}(b). with the decrease of $\alpha$ from 0.04 to 0.20, the general variation trend of the mean accuracy curve is negative, i.e., smaller $\alpha$ can yield higher accuracy. When $\alpha$ is 0.04, the accuracy reaches the maximum 94.45\%. The reason that we did not reduce $\alpha$ since 0.04 is that the DGE is decreasingly effective when $\alpha$ becomes lower since 0.04. We draw this conclusion from two observations: 1) when we trained AH-Net (when $\alpha$ is 0.02 or 0.03) for many times, the accuracy sometimes is lower than 10\% (\textit{i.e.}, dommain gap eliminator is sometimes ineffective), 2) when $\alpha$ is 0.01, the accuracy is always lower than 10\% (\textit{i.e.}, the DGE is always ineffective). In brief, DI can bring a recognition accuracy increase of $80\%+$ on CNN.

\subsection{Effect of $\alpha$ on KNN}
Similar to the experiments on CNN, we varied $\alpha$ from 0.05 to 0.25 with a stride of 0.01. The experiment results shown in Fig. \ref{fig:performance2}(a). The general variation trend of the accuracy is: with the increase of $\alpha$ before 0.19, the accuracy is increasing, afterwards the accurac is decreasing with the increase of $\alpha$. When $\alpha$, the accuracy curve reaches its peak 87.33\%. This accuracy outperforms EI. 

\subsection{Effect of $\alpha$ on SVM}
Similar to the experiment on KNN, we show the results in Fig. \ref{fig:performance2}(b). The variation trend of the SVM accuracy is similar to the KNN accuracy. SVM achieves the best accuracy 90,12\% when $\alpha$ is 0.19.

\begin{table}
\begin{center}
{\caption{Comparing DI with EI, CrossSense, WiAG and Widar3.0.}\label{tab:comparison}}
\begin{tabular}{ccccccc}
\cline{1-6}
\rule{0pt}{12pt}
Approach&DI&EI&CrossSense&WiAG&Widar3.0\\
\hline
\\[-6pt]
\quad Accuracy&94.45\%&$\le90\%$&$90\%+$&91.40\%&92.70\%\\
\quad S. G. P.&Yes&Yes&Yes&Yes&No\\
\quad B M. L. T.&Yes&Yes&No&Yes&Yes\\
\quad G. P. C.&No&Yes&No&No&No\\
\hline
\end{tabular}
\end{center}
\end{table}

\subsection{Comparison with existing works}
We compared DI with four state-of-the-art approaches: EI, CrossSense, WiAG and Widar3.0. The comparison results are shown in Table \ref{tab:comparison}. S. G. P. in the first column means that static gesture is permitted. B. M. L. T. means that the number of basic learning models is less than three. G. P. C. means the gesture percentage constraint. As one can see from Table \ref{tab:comparison}, DI achieve the highest accuracy and the least recognition constraints.

\section{Conclusion}
This paper first presented the concept of domain gap in WiFi-based gesture recognition and then proposed a framework DI to eliminate domain gap. DI employs a novel deep model, namely AH-Net, to generate domain gap eliminator, further to achieve domain-independent gesture recognition. The experiments on ten domains and ten gestures show that DI can achieve the recognition accuracy of 94.45\%, which is better than previous works.

\bibliographystyle{IEEEtran}
\bibliography{IEEEexample}

\begin{thebibliography}{10}
\providecommand{\url}[1]{#1}
\csname url@samestyle\endcsname
\providecommand{\newblock}{\relax}
\providecommand{\bibinfo}[2]{#2}
\providecommand{\BIBentrySTDinterwordspacing}{\spaceskip=0pt\relax}
\providecommand{\BIBentryALTinterwordstretchfactor}{4}
\providecommand{\BIBentryALTinterwordspacing}{\spaceskip=\fontdimen2\font plus
\BIBentryALTinterwordstretchfactor\fontdimen3\font minus
  \fontdimen4\font\relax}
\providecommand{\BIBforeignlanguage}[2]{{%
\expandafter\ifx\csname l@#1\endcsname\relax
\typeout{** WARNING: IEEEtran.bst: No hyphenation pattern has been}%
\typeout{** loaded for the language `#1'. Using the pattern for}%
\typeout{** the default language instead.}%
\else
\language=\csname l@#1\endcsname
\fi
#2}}
\providecommand{\BIBdecl}{\relax}
\BIBdecl

\bibitem{DBLP:conf/cvpr/GkioxariGDH18}
G.~Gkioxari, R.~B. Girshick, P.~Doll{\'{a}}r, and K.~He, ``Detecting and
  recognizing human-object interactions,'' in \emph{{IEEE} Conference on
  Computer Vision and Pattern Recognition, {CVPR}}, 2018, pp. 8359--8367.

\bibitem{DBLP:journals/sigmobile/LiLZ16}
T.~Li, Q.~Liu, and X.~Zhou, ``Practical human sensing in the light,''
  \emph{GetMobile Mob. Comput. Commun.}, vol.~20, no.~4, pp. 28--33, 2016.

\bibitem{DBLP:conf/icassp/KalgaonkarR09}
K.~Kalgaonkar and B.~Raj, ``One-handed gesture recognition using ultrasonic
  doppler sonar,'' in \emph{Proceedings of the {IEEE} International Conference
  on Acoustics, Speech, and Signal Processing, {ICASSP}}, 2009, pp. 1889--1892.

\bibitem{DBLP:journals/imwut/NandakumarTKG17}
R.~Nandakumar, A.~Takakuwa, T.~Kohno, and S.~Gollakota, ``Covertband: Activity
  information leakage using music,'' \emph{{IMWUT}}, vol.~1, no.~3, pp.
  87:1--87:24, 2017.

\bibitem{DBLP:journals/imwut/GuanP17}
Y.~Guan and T.~Pl{\"{o}}tz, ``Ensembles of deep {LSTM} learners for activity
  recognition using wearables,'' \emph{{IMWUT}}, vol.~1, no.~2, pp.
  11:1--11:28, 2017.

\bibitem{DBLP:journals/csur/BullingBS14}
A.~Bulling, U.~Blanke, and B.~Schiele, ``A tutorial on human activity
  recognition using body-worn inertial sensors,'' \emph{{ACM} Comput. Surv.},
  vol.~46, no.~3, pp. 33:1--33:33, 2014.

\bibitem{DBLP:conf/mobicom/WangLCG0L14}
Y.~Wang, J.~Liu, Y.~Chen, M.~Gruteser, J.~Yang, and H.~Liu, ``E-eyes:
  device-free location-oriented activity identification using fine-grained wifi
  signatures,'' in \emph{The 20th Annual International Conference on Mobile
  Computing and Networking, MobiCom'14}, 2014, pp. 617--628.

\bibitem{DBLP:conf/infocom/Abdel-NasserYH15}
H.~Abdelnasser, M.~Youssef, and K.~A. Harras, ``Wigest: {A} ubiquitous
  wifi-based gesture recognition system,'' in \emph{2015 {IEEE} Conference on
  Computer Communications, {INFOCOM}}, 2015, pp. 1472--1480.

\bibitem{DBLP:conf/mobisys/VenkatnarayanPS18}
R.~H. Venkatnarayan, G.~Page, and M.~Shahzad, ``Multi-user gesture recognition
  using wifi,'' in \emph{Proceedings of the 16th Annual International
  Conference on Mobile Systems, Applications, and Services, MobiSys}, 2018, pp.
  401--413.

\bibitem{wang2018csi}
F.~Wang, J.~Han, S.~Zhang, X.~He, and D.~Huang, ``Csi-net: Unified human body
  characterization and pose recognition,'' \emph{arXiv preprint
  arXiv:1810.03064}, 2018.

\bibitem{DBLP:conf/mobicom/JiangMMYWYXSMKX18}
W.~Jiang, C.~Miao, F.~Ma, S.~Yao, Y.~Wang, Y.~Yuan, H.~Xue, C.~Song, X.~Ma,
  D.~Koutsonikolas, W.~Xu, and L.~Su, ``Towards environment independent device
  free human activity recognition,'' in \emph{Proceedings of the 24th Annual
  International Conference on Mobile Computing and Networking, MobiCom}, 2018,
  pp. 289--304.

\bibitem{DBLP:journals/cea/RighiGKDC20}
R.~da~Rosa~Righi, G.~Goldschmidt, R.~Kunst, C.~Deon, and C.~A. da~Costa,
  ``Towards combining data prediction and internet of things to manage milk
  production on dairy cows,'' \emph{Comput. Electron. Agric.}, vol. 169, p.
  105156, 2020.

\bibitem{DBLP:conf/mobisys/VirmaniS17}
A.~Virmani and M.~Shahzad, ``Position and orientation agnostic gesture
  recognition using wifi,'' in \emph{Proceedings of the 15th Annual
  International Conference on Mobile Systems, Applications, and Services,
  MobiSys'17}, 2017, pp. 252--264.

\bibitem{DBLP:conf/mobisys/ZhengZ0ZLW019}
Y.~Zheng, Y.~Zhang, K.~Qian, G.~Zhang, Y.~Liu, C.~Wu, and Z.~Yang,
  ``Zero-effort cross-domain gesture recognition with wi-fi,'' in
  \emph{Proceedings of the 17th Annual International Conference on Mobile
  Systems, Applications, and Services, MobiSys}, 2019, pp. 313--325.

\bibitem{DBLP:conf/iccv/0037ZPHH19}
F.~Wang, S.~Zhou, S.~Panev, J.~Han, and D.~Huang, ``Person-in-wifi:
  Fine-grained person perception using wifi,'' in \emph{2019 {IEEE/CVF}
  International Conference on Computer Vision, {ICCV}}, 2019, pp. 5451--5460.

\bibitem{DBLP:conf/infocom/WangLCLXWHL18}
C.~Wang, J.~Liu, Y.~Chen, H.~Liu, L.~Xie, W.~Wang, B.~He, and S.~Lu, ``Multi -
  touch in the air: Device-free finger tracking and gesture recognition via
  {COTS} {RFID},'' in \emph{{IEEE} Conference on Computer Communications,
  {INFOCOM}}, 2018, pp. 1691--1699.

\bibitem{DBLP:journals/jsac/WangLSLL17}
W.~Wang, A.~X. Liu, M.~Shahzad, K.~Ling, and S.~Lu, ``Device-free human
  activity recognition using commercial wifi devices,'' \emph{{IEEE} Journal on
  Selected Areas in Communications}, vol.~35, no.~5, pp. 1118--1131, 2017.

\bibitem{DBLP:conf/mobicom/PuGGP13}
Q.~Pu, S.~Gupta, S.~Gollakota, and S.~Patel, ``Whole-home gesture recognition
  using wireless signals,'' in \emph{The 19th Annual International Conference
  on Mobile Computing and Networking, MobiCom'13}, 2013, pp. 27--38.

\bibitem{DBLP:journals/imwut/HanQYWDLR18}
J.~Han, C.~Qian, Y.~Yang, G.~Wang, H.~Ding, X.~Li, and K.~Ren, ``Butterfly:
  Environment-independent physical-layer authentication for passive {RFID},''
  \emph{{IMWUT}}, vol.~2, no.~4, pp. 166:1--166:21, 2018.

\bibitem{DBLP:journals/gandc/HorrocksHWWF19}
T.~Horrocks, E.~Holden, D.~Wedge, C.~Wijns, and M.~Fiorentini, ``Geochemical
  characterisation of rock hydration processes using t-sne,'' \emph{Comput.
  Geosci.}, vol. 124, pp. 46--57, 2019.

\bibitem{DBLP:journals/corr/GoodfellowSS14}
I.~J. Goodfellow, J.~Shlens, and C.~Szegedy, ``Explaining and harnessing
  adversarial examples,'' in \emph{3rd International Conference on Learning
  Representations, {ICLR}}, 2015.

\bibitem{DBLP:journals/ccr/HalperinHSW11}
D.~Halperin, W.~Hu, A.~Sheth, and D.~Wetherall, ``Tool release: gathering
  802.11n traces with channel state information,'' \emph{Computer Communication
  Review}, vol.~41, no.~1, p.~53, 2011.

\bibitem{DBLP:journals/sensors/KangSZZG20}
Y.~Kang, Z.~Shi, H.~Zhang, D.~Zhen, and F.~Gu, ``A novel method for the dynamic
  coefficients identification of journal bearings using kalman filter,''
  \emph{Sensors}, vol.~20, no.~2, p. 565, 2020.

\end{thebibliography}
\end{document}